\documentclass[11pt]{article}

\usepackage[final]{acl}
\usepackage{stroop}

\usepackage{times}
\usepackage{latexsym}

\usepackage[T1]{fontenc}

\usepackage[utf8]{inputenc}
\usepackage{arydshln}

\usepackage{microtype}

\usepackage{inconsolata}

\usepackage{graphicx}

\usepackage{framed}
\usepackage{fvextra}
\usepackage{tcolorbox}
\usepackage{xcolor}
\usepackage{fvextra}

\newtcolorbox{orangeprompt}{
  colback=orange!5,
  colframe=orange!70!black,
  boxrule=1.2pt,
  arc=0pt,
  left=1mm,
  right=1mm,
  top=1mm,
  bottom=1mm
}

\newtcolorbox{greenprompt}{
  colback=green!5,
  colframe=green!70!black,
  boxrule=1.2pt,
  arc=0pt,
  left=1mm,
  right=1mm,
  top=1mm,
  bottom=1mm
}

\newtcolorbox{redprompt}{
  colback=pink!3,
  colframe=pink!70!black,
  boxrule=1.2pt,
  arc=0pt,
  left=1mm,
  right=1mm,
  top=1mm,
  bottom=1mm
}

\title{UvA-MT’s Participation in the WMT26 General Translation Shared Task}
\author{
    {\bf Di Wu$^{*}$ \hspace{2mm}}
    {\bf Sergey Troshin$^{*}$ \hspace{2mm}}
    {\bf Wafaa Mohammed \hspace{2mm}}
    {\bf Seth Aycock \hspace{2mm}} \\
    {\bf Evgeniia Tokarchuk \hspace{2mm}} 
    {\bf Vlad Niculae \hspace{2mm}}
    {\bf Christof Monz\hspace{2mm}}\\
    Language Technology Lab, 
    University of Amsterdam\\ 
    \texttt{\{d.wu, s.troshin, w.m.a.mohammed, s.aycock\}@uva.nl} \\
    \texttt{\{e.tokarchuk, v.niculae, c.monz\}@uva.nl}
 }

\begin{document}
\maketitle
\begin{abstract}
This paper presents UvA-MT's submission to the WMT 2026 General Machine Translation shared task, competing in the unconstrained track across all 23 translation directions.
This year, we fully leverage the test-time methods of Large Language Models (LLMs) for machine translation. Specifically, (1) we use sequential sampling with a refinement prompt to generate a pool of translation candidates for each source segment, a strategy shown to outperform traditional parallel sampling~\citep{wu2026ladders}; and (2) we employ LLMs as pairwise judges to select the best candidate via a round-robin voting mechanism, with offline evaluation on the WMT25 benchmark showing a \textbf{very clear} improvement over point-wise best-of-$N$ selection.
Combined, these two strategies form our WMT26 submission.

\end{abstract}

\section{Introduction}
\def\thefootnote{*}\footnotetext{Equal first-author contribution.}\def\thefootnote{\arabic{footnote}}
In this paper, we describe our submission to the WMT 2026 General Machine Translation shared task (unconstrained track), which covers 23 translation directions. Motivated by recent advances in Large Language Models (LLMs), particularly test-time scaling techniques \citep{tan2025investigating,wu2025please,wu2026ladders}, we investigate to what extent translation quality can be improved by relying solely on general-purpose LLMs at inference time.

\begin{figure*}
    \centering
    \includegraphics[width=0.8\linewidth]{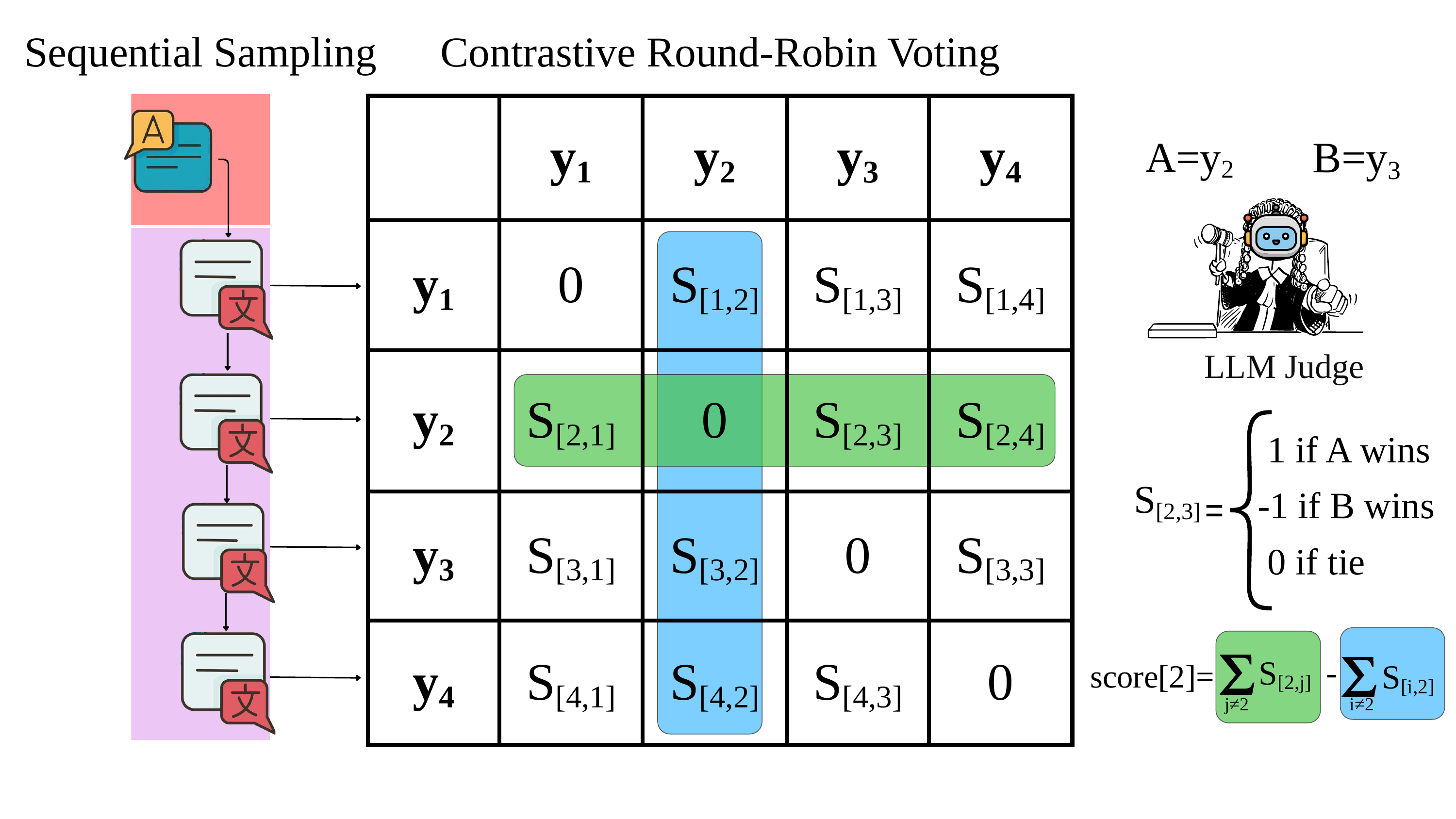}
    \caption{An overview of our pipeline: (left) sequential sampling to generate a list of candidates, see Section~\ref{sec-2.1} for details; (right) contrastive round-robin voting with a LLM-as-a-Judge to select the best sample. Note that, to eliminate positional bias in LLM-based pairwise judging, each pair of samples is evaluated twice with their order swapped. The final score is computed by combining the scores that a hypothesis achieves in both directions, see Section~\ref{sec-2.2} for details.}
    \label{fig:judge_image}
\end{figure*}

\Cref{fig:judge_image} illustrates an overview of our approach. Our submission uses two test-time technologies: 
\begin{itemize}
    \item \textbf{Sequential Sampling:} Given a source input to a LLM, a pool of translations is sequentially sampled, with each subsequent translation instructed to improve upon its predecessor. This recursive sampling strategy exploits the LLM's self-improvement capability and has been shown by \citet{wu2026ladders} to be more effective than traditional parallel sampling, especially under limited inference budgets.
    
    \item \textbf{Contrastive Round-Robin Voting:} Given a pool of translation candidates, every pair is compared head-to-head by a LLM judge that explicitly contrasts their differences (rather than scoring each independently). Wins are tallied across all pairwise match-ups, and the candidate with the most victories is selected as the final output. We argue that this strategy is more reliable than traditional best-of-$N$ sampling because pairwise judgments require only relative preferences rather than absolute quality scores, reducing the need for calibration and enabling finer-grained discrimination among candidates~\citep{ouyang2022training}. 
    
\end{itemize}

Section~\ref{sec-2} describes our main methodology, which consists of two components: 1) sequential sampling, i.e., how to generate a pool of translation candidates for each source segment; and 2) contrastive round-robin voting with a pairwise LLM judge, i.e., how to select the best candidate from the resulting pool.

In Section~\ref{sec-3.1}, we conduct offline evaluations on the WMT24 benchmark~\citep{kocmi-etal-2024-findings,deutsch-etal-2025-wmt24}, following \citet{wu2026ladders}, showing that \textbf{sequential sampling} yields substantial translation improvements over parallel sampling, especially under limited sampling budgets. 

In Section~\ref{sec-3.2}, we show that the \textbf{contrastive round-robin voting} mechanism, powered by a LLM-based pairwise (contrastive) judge, consistently outperforms the traditional best-of-$N$ sampling paradigm with a pointwise judge (metric model, e.g., CometKiwi). This is reflected in both improved Best@$N$ performance and stronger rank correlation, as measured by Kendall's $\tau$ and Spearman's $\rho$, on the WMT25 metrics benchmark~\citep{lavie-etal-2025-findings}.

Lastly, we apply these two strategies for our WMT26 submission, see details in Section~\ref{sec-4}. 

\section{Methodology}
\label{sec-2}
\subsection{Generation of $N$ candidates}
\label{sec-2.1}
Our generation setting follows the previous and concurrent line of work, which we refer to as \textit{sequential} sampling \citep{wu2025please, wu2026ladders}. The key idea of this sampling strategy is to refine the previously generated hypotheses, instead of generating fully in parallel, which demonstrated higher sampling efficiency given a smaller sampling budget. In particular, to obtain a set of candidates for reranking, we translate each document sequentially $8$ times, including the initial translation and $7$ refinement rounds.
Formally, for a sampling turn $t$, \textit{sequential} sampling generates a translation $y$ as follows:
\begin{equation}
    y^{(t)} \sim p_\theta\left(\cdot \mid I^{(r)},y^{(t-1)}, I(x)\right),
\end{equation}
\noindent where $I^{(r)}$ is a prompt requesting a model to revisit a translation to improve it \citep{wu2025please}; and $I(x)$ includes a system prompt and the initial turn translation request. To reduce the computational cost, here we use the $\text{history}-1$ formulation of the sequential sampling, shown to be most efficient and effective in \citet{wu2026ladders}, feeding only the most recent previous translation to the model, thus capping the context size growth.

We use the following prompt template for the candidate generation step:
\begin{orangeprompt}
\begin{Verbatim}[breaklines=true, 
                 breakanywhere=true,
                 breaksymbolright={},
                 breaksymbolsepleft=0pt,
                 breaksymbolindent=0pt,
                 breaksymbolleft={},
                 fontsize=\small]

[system]
You are a professional translator. Follow the instruction and output only the translation, with no explanations.

Task-specific instruction:
{instruction}
\end{Verbatim}
\end{orangeprompt}
\begin{greenprompt}
\begin{Verbatim}[breaklines=true, 
                 breakanywhere=true,
                 breaksymbolright={},
                 breaksymbolsepleft=0pt,
                 breaksymbolindent=0pt,
                 breaksymbolleft={},
                 fontsize=\small]
[user turn initial]
Please translate the following text from {src_lang} to {tgt_lang}. Provide only one translation on the first line and do not output anything else after that.

{src_lang}: {source}
{tgt_lang}:

[assistant turn]
{assistant response}
\end{Verbatim}
\end{greenprompt}
\begin{redprompt}
\begin{Verbatim}[
  breaklines=true,
  breakanywhere=true,
  breaksymbolright={},
  breaksymbolsepleft=0pt,
  breaksymbolindent=0pt,
  breaksymbolleft={},
  fontsize=\small
]

[user turn revisiting]
Please again translate the following text from {src_lang} to {tgt_lang} to make it better. Provide only one translation on the first line and do not output anything else after that.

{src_lang}: {source}
{tgt_lang}:

[assistant turn]
{assistant response}

\end{Verbatim}
\end{redprompt}

\subsection{Contrastive Judge}
\label{sec-2.2}
We use the following prompt template for the pairwise judges:

\begin{orangeprompt}
\begin{Verbatim}[breaklines=true, 
                 breakanywhere=true,
                 breaksymbolright={},
                 breaksymbolsepleft=0pt,
                 breaksymbolindent=0pt,
                 breaksymbolleft={},
                 fontsize=\small]

[system]
You are an expert, impartial bilingual translation evaluator. You compare two candidate translations of the same source and decide which is better. Judge strictly on quality; ignore which candidate is labelled A or B and ignore their order or length.

\end{Verbatim}
\end{orangeprompt}
\begin{greenprompt}
\begin{Verbatim}[breaklines=true, 
                 breakanywhere=true,
                 breaksymbolright={},
                 breaksymbolsepleft=0pt,
                 breaksymbolindent=0pt,
                 breaksymbolleft={},
                 fontsize=\small]
[user]
Compare two candidate translations and decide which is better.

Source language: {src_lang}
Target language: {tgt_lang}

Judge against this rubric (earlier criteria matter more):
Pick the translation that is more accurate and more natural in the target language.

{instruction_block}=== SOURCE ===
{source}

=== TRANSLATION A ===
{trans_a}

=== TRANSLATION B ===
{trans_b}

Return only a single JSON object on the last line and nothing after it:
{"winner": "A" | "B" | "tie"}

[assistant]
{assistant response}

\end{Verbatim}
\end{greenprompt}
The `\texttt{\{instruction\_block\}}' here is populated when the record carries a non-trivial instruction. Then it resolves to `\texttt{The translations were produced under this task-specific instruction; also judge how faithfully each candidate follows it:\textbackslash n\{instruction \}}`. Trivial ``\texttt{Translate from X to Y.}" instructions are dropped, as we incorporate the source and target language information directly into the user prompt. We use the constrained generation for json, implemented on the server side, whereas we provide the json schema.

We use temperature $0$ for the judge, and we choose the lowest possible reasoning effort for each model.

\begin{figure}[t]
    \centering
    \includegraphics[width=0.95\linewidth]{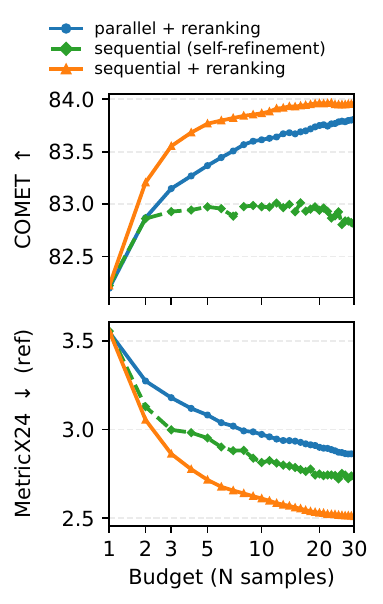}
    \caption{Sequential + reranking demonstrates  superior quality@budget trade-off, compared to the parallel sampling + reranking, and sequential sampling (self-refinement), following \citet{wu2026ladders}. Here we report the quality of the methods on WMT24++ with the \textit{gpt-4o-mini-2024-07-18}; metrics aggregated over the 6 languages.}
    \label{fig:seqpar}
\end{figure}

\subsection{Contrastive Round-Robin Voting}
The idea of a contrastive round-robin voting manifests itself as early as in the works of \citet{llull1299}~\footnote{assembled and translated by \citet{hagele2001llull}}, where they propose to evaluate the candidates during election via full round-robin contrastive voting, such that a candidate who is victorious in most of the pairwise comparisons is named the winner (also known as Copeland's method). 

In our formulation, the contrastive round-robin voting with LLMs-as-a-judge consists of the pairwise judge scoring and the score aggregation. By solely relying on contrastive LLMs-as-a-judge prompt templates to obtain pairwise votes, we use the following counting aggregation strategy for our robust evaluation. Formally, given a pool of $N$ candidates, we define a $N \times N$ score matrix $S$, where $S[i, j] \in \{1, -1, 0\}$ denotes a voting result for a candidate $i$ agains a candidate $j$, where the verdict $1$ stands for a win, $-1$ for a loss, and $0$ for a tie. Then a ranking score of each candidate is a sum of its points: 
$
    \text{score}[i]=\sum_{j} S[i,j].
$
LLM judges are known to have strong position biases \citep{wang-etal-2024-large-language-models-fair}, i.e., they can be inconsistent when changing a candidate order. To mitigate the position bias, we evaluate both $i$ vs $j$ and $j$ vs $i$ directions (the order of these is swapped in the judge prompt template). The de-biased score we use is as follows:
\begin{equation}
    \text{score}[i]=\sum_{j=1}^{N} S[i,j] - \sum_{j=1}^{N} S[j,i].
\end{equation}

\section{Offline Evaluation}

\subsection{On the Effectiveness of Sequential Sampling}
\label{sec-3.1}
We choose sequential sampling as the main generation strategy to form a pool of samples for subsequent reranking. Here, we briefly report an abridged version of the results from \citet{wu2026ladders} in \Cref{fig:seqpar}, demonstrating the comparison of parallel and sequential strategy for the \texttt{GPT-4o-mini}\footnote{gpt-4o-mini-2024-07-18} model. 

Here, we compare \textbf{parallel + reranking}, \textbf{sequential (self-refinement)}  and \textbf{sequential + reranking}. These results report the average machine translation quality on WMT24++, aggregated over $6$ target languages: Dutch, German, Mandarin, Standard Arabic, Russian, Romanian; measured using a reference-based \href{https://huggingface.co/Unbabel/wmt22-comet-da}{COMET}~\citep{rei-etal-2022-comet}, and reference-based \href{https://huggingface.co/google/metricx-24-hybrid-xxl-v2p6-bfloat16}{MetricX-24-Hybrid-XXL (ref)}~\citep{juraska-etal-2024-metricx}. Here, \href{https://huggingface.co/Unbabel/wmt23-cometkiwi-da-xl}{Comet-Kiwi-XL}~\citep{rei-etal-2023-scaling} is used as a metric for reranking, following  \citet{wu2026ladders}. The results (\Cref{fig:seqpar}) show that the combination of \textbf{sequential sampling with reranking} outperforms both the parallel sampling with reranking and simple self-refinement, motivating our choice for the sampling method. 

In the next section (\Cref{sec-3.2}), we revisit the reranking step, the main focus of this submission.

\begin{table*}[t]

\centering

\small

\begin{tabular}{l l l c c c c c c c}

\hline
\hline

Rank & Type & System & $\tau_a$ & $\rho$ & Best@1 & Best@2 & Tie & Disagree \\

\hline

1  & judge  & Gemini-3.5-Flash (reasoning) &  \textbf{23.1} &           30.9 &           17.7 &           32.8 &  \textbf{12.8} &  \textbf{12.6} \\

2  & judge  & GPT-5.4 (reasoning)          &           22.6 &           30.6 &  \textbf{22.2} &  \textbf{36.9} &           17.2 &           16.9 \\

3  & judge  & Gemini-3.5-Flash (direct)    &           22.4 &  \textbf{31.1} &           20.2 &           33.8 &           13.9 &           13.5 \\

4  & judge  & GPT-5.4 (direct)             &           20.8 &           28.7 &           19.7 &           32.3 &           18.4 &           18.0 \\

5  & judge  & GPT-4o-mini (direct)         &           16.7 &           25.9 &           21.7 &           33.8 &           31.6 &           31.3 \\

6  & judge  & GPT-4o-mini (reasoning)      &           12.4 &           23.7 &           13.6 &           28.8 &           53.7 &           53.5 \\

7  & judge  & Gemini-2.5-Flash (reasoning) &           10.3 &           26.5 &           21.2 &  \textbf{36.9} &           62.1 &           47.7 \\

8  & judge  & Gemini-2.5-Flash (direct)    &           10.0 &           23.8 &           20.7 &           33.8 &           59.3 &           51.4 \\

\hdashline

9  & metric & COMET-Kiwi-23-XXL            &            5.2 &            6.9 &           14.1 &           25.8 &            0.3 &            0.0 \\

10 & metric & MetricX-24-Hyb-XL            &            4.9 &            6.8 &           17.2 &           29.3 &            0.3 &            0.0 \\

11 & metric & MetricX-24-Hyb-XXL           &            2.3 &            3.3 &           18.2 &           29.8 &            0.3 &            0.0 \\

12 & metric & COMET-Kiwi-22                &            1.6 &            2.6 &           15.2 &           25.8 &            0.3 &            0.0 \\

13 & metric & COMET-Kiwi-23-XL             &            1.6 &            2.6 &           15.7 &           25.3 &            0.3 &            0.0 \\

14 & metric & MetricX-23-QE-L              &           -2.8 &           -4.4 &           15.7 &           27.8 &            0.3 &            0.0 \\

\hline
\hline

\end{tabular}

\caption{EN-ZH results ($n=198$ documents), reported as percentages. $\tau_a$ is the per-pair-verdict Kendall, which is the coefficient comparable across the judge/metric boundary and is used for system ranking. The best value among judge models is shown in bold. For Tie and Disagree, lower is better.}

\label{tab:judge_metric_enzh}
\end{table*}

\begin{table*}[t]

\centering

\small

\begin{tabular}{l l l c c c c c c}

\hline
\hline

Rank & Type & System & $\tau_a$ & $\rho$ & Best@1 & Best@2 & Tie & Disagree \\

\hline

1  & judge  & GPT-5.4 (direct)             &  \textbf{30.8} &  \textbf{42.5} &           36.0 &           51.5 &           17.2 &           16.8 \\

2  & judge  & Gemini-3.5-Flash (reasoning) &           30.8 &           41.7 &  \textbf{37.0} &           52.0 &           11.9 &           11.6 \\

3  & judge  & Gemini-3.5-Flash (direct)    &           30.6 &           41.6 &           36.0 &           52.0 &  \textbf{10.6} &  \textbf{10.3} \\

4  & judge  & GPT-5.4 (reasoning)          &           29.7 &           40.1 &           35.5 &  \textbf{54.5} &           16.8 &           16.5 \\

5  & judge  & GPT-4o-mini (direct)         &           22.4 &           34.2 &           29.5 &           41.0 &           30.5 &           30.2 \\

6  & judge  & GPT-4o-mini (reasoning)      &           19.8 &           33.4 &           28.0 &           43.0 &           43.7 &           43.4 \\

7  & judge  & Gemini-2.5-Flash (reasoning) &           16.0 &           38.3 &           32.0 &           47.5 &           53.2 &           45.0 \\

8  & judge  & Gemini-2.5-Flash (direct)    &           15.7 &           35.9 &           32.5 &           48.0 &           54.3 &           48.3 \\

\hdashline

9  & metric & COMET-Kiwi-23-XXL            &            8.6 &           11.4 &           17.0 &           28.5 &            0.3 &            0.0 \\

10 & metric & MetricX-24-Hyb-XXL           &            6.4 &            8.1 &           17.5 &           28.5 &            0.7 &            0.0 \\

11 & metric & MetricX-24-Hyb-XL            &            6.3 &            7.5 &           13.0 &           22.5 &            0.3 &            0.0 \\

12 & metric & COMET-Kiwi-23-XL             &            4.9 &            6.1 &           16.0 &           28.5 &            0.3 &            0.0 \\

13 & metric & COMET-Kiwi-22                &            2.2 &            2.7 &           11.0 &           23.0 &            0.3 &            0.0 \\

14 & metric & MetricX-23-QE-L              &            0.2 &           -0.3 &           16.5 &           25.5 &            0.3 &            0.0 \\

\hline
\hline

\end{tabular}

\caption{EN-RU results ($n=200$ documents), reported as percentages. $\tau_a$ is the per-pair-verdict Kendall, which is the coefficient comparable across the judge--metric boundary and is used for system ranking. The best value among judge models is shown in bold. For Tie and Disagree, lower is better.}

\label{tab:judge_metric_enru}

\end{table*}

\subsection{On the Effectiveness of Contrastive Round-Robin Voting}
\label{sec-3.2}

Given a pool of hypotheses for each source text, we use Round-Robin Voting with commercial LLMs to select the winning hypothesis for the final submission. 
To show the effectiveness, we compare our strategy with traditional Best-of-$N$ sampling, known as Quality-aware decoding in MT \citep{fernandes-etal-2022-quality}, which uses machine translation quality estimation models for point-wise ranking, e.g., MetricX-QE or Comet-Kiwi.

We use the WMT25 General MT Human Evaluation dataset\footnote{https://github.com/wmt-conference/wmt25-general-mt} as the benchmark. We evaluate two translation directions, EN$\rightarrow$ZH and EN$\rightarrow$RU, which contain 198 and 200 paragraph-level source texts, respectively. Each source text is accompanied by translations produced by at most $18$ MT systems, including last year's leading systems such as GPT-4.1, Claude-4, UvA-MT, and Qwen3-235B~\citep{achiam2023gpt,wu-etal-2025-uva,yang2025qwen3,kocmi-etal-2025-findings}.

For each source text in the dataset, we randomly sample 8 translations. We retain the corresponding human evaluation Error Span Annotation \citep[ESA]{kocmi-etal-2024-error} scores for each translation (an ordinal score from $0$ to $100$). For translations annotated by multiple experts, we average the individual scores to obtain the final human evaluation score.

We evaluate pairwise judgments using contrastive Round-Robin Voting across different LLMs and compare them with traditional machine translation (MT) evaluation metrics by measuring their Kendall rank correlation ($\tau_{a}$) with human judgments.
The resulting rankings are presented in Table~\ref{tab:judge_metric_enzh} and Table~\ref{tab:judge_metric_enru}. Note that Kendall rank correlation is comparable across all of these systems. Additionally, we report Spearman's rank correlation coefficient ($\rho$), Best@1, Best@2, rate of a tie, and disagreement rate.

We find that:

\begin{itemize}
    \item Traditional MT metrics \textbf{completely fail} to rank last year's translation hypotheses. Even the best-performing metric, COMET-Kiwi-23-XXL, achieves Kendall's $\tau_a$ scores of only 5.2 and 8.6 in EN-ZH and EN-RU. Its Best@1 and Best@2 accuracies are approximately 12.5\% and 25\%, respectively, suggesting performance that is close to random guessing on this ranking task.
    \item Our pairwise LLM-as-a-Judge framework with Round-Robin Voting \textbf{substantially outperforms} all point-wise translation metrics, achieving consistent and very clear improvements in Kendall's $\tau_{a}$ and Spearman's $\rho$, as well as Best@1 and Best@2 accuracy.
    \item Strong commercial LLMs, such as GPT-5.4 and Gemini-3.5-Flash, perform substantially better, exhibiting significantly lower tie rates. In contrast, weaker models, such as Gemini-2.5-Flash, produce a much higher tie rate, suggesting a limited ability to discriminate among high-quality translation candidates.
    \item Pairwise LLM judges exhibit position bias (Disagree): reversing the order of a translation pair can cause the preferred candidate to switch. However, strong LLMs, e.g., Gemini-3.5-Flash, perform much better than weaker models here, e.g., in Chinese, Gemini-3.5-Flash has a score of 12.6 while Gemini-2.5-Flash has a score of 51.4.
    \item Reasoning sometimes helps judging, but the gains are marginal compared with direct judging. For example, Gemini-3.5-Flash with reasoning only slightly outperforms its direct-judge counterpart (23.1 vs.\ 22.4 in $\tau_a$), while even achieving lower Spearman's $\rho$ and Best@1 accuracy.
\end{itemize}

Considering its strong performance and low cost, we selected \texttt{gemini-3.5-flash (direct)} as the judge model for our final submission.

\section{Final Ensemble}
\label{sec-4}
For the final submission, we run the generation using \texttt{gemini-3.5-flash} and \texttt{GPT-5.4}, selecting a winning candidate from each family using \texttt{gemini-3.5-flash (direct)}  as a judge. 
\paragraph{Ensembling 2 model families.} For two models (a, b), let us denote $S_{a,a}, S_{b,b}$ as score matrices for intra-model evaluation, and $S_{a,b}, S_{b,a}$ as inter-model evaluation. We do not fully compute $S_{a,b}, S_{b,a}$, but only compute the cross-model scores for the selected winners ($w_a$ and $w_b$):
\begin{equation}
\begin{split}
    \text{score}[w_a] ={} & \sum_{j\ne w_a} S_{a,a}[w_a,j] - \sum_{j\ne w_a} S_{a,a}[j,w_a] \\
    & + \sum_{j} S_{a,b}[w_a,j] - \sum_{j} S_{b,a}[j,w_a].
\end{split}
\end{equation}
\begin{equation}
\begin{split}
    \text{score}[w_b] ={} & \sum_{j\ne w_b} S_{b,b}[w_b,j] - \sum_{j\ne w_b} S_{b,b}[j,w_b] \\
    & + \sum_{j} S_{b,a}[w_b,j] - \sum_{j} S_{a,b}[j,w_b].
\end{split}
\end{equation}

We run $2N(N-1)$ judge calls within the groups and $4N$ cross-group judge calls for the winners (assuming all groups are of size $N$), in total we run $2N(N-1) + 4N - 2 = 142$ calls. A full $2N=16$ round robin tournament would require $2N(2N-1)=240$ judge calls. Although other efficient tournament variants are possible, we use this strategy as a simple and easy to scale hierarchical approximation to the full round-robin tournament.

\section{Conclusion}

In sum, we implement a combination of sequential candidate sampling and contrastive round-robin voting for reranking. Our contrastive reranking strategy demonstrates  much higher correlation with human judgements than more standard quality-aware decoding techniques.

\section{Acknowledgments}
This work is supported in part by project VI.Veni.212.228 of the research program
`Veni', which is financed by the Dutch Research Council (NWO); this research is also funded in part by the Netherlands Organization for Scientific Research (NWO) under project numbers VI.C.192.080 and 2023.017; and is part of
‘Hybrid Intelligence: augmenting human intellect’
(https://hybrid-intelligence-centre.nl) with project number 024.004.022 of the
research program `Gravitation' which is (partly) financed by the Dutch Research
Council (NWO).

\bibliography{anthology1,anthology2,custom}

\end{document}